\documentclass[runningheads]{llncs}
\usepackage[T1]{fontenc}
\usepackage{graphicx}

\usepackage{amsmath,amssymb}

\usepackage{tabularx}
\usepackage{booktabs}
\usepackage{xcolor}
\usepackage{bm}
\usepackage{caption}

\usepackage{url}

\usepackage[colorlinks=true,linkcolor=blue, citecolor=blue]{hyperref}
\usepackage{xcolor}
\definecolor{newcolor}{rgb}{.8,.349,.1}

\let\llncssubparagraph\subparagraph
\let\subparagraph\paragraph
\usepackage[compact]{titlesec}
\titlespacing{\section}{3pt}{3pt}{3pt}
\AtBeginDocument{
    \setlength\abovedisplayskip{3pt}
    \setlength\belowdisplayskip{3pt}}
\let\subparagraph\llncssubparagraph

\usepackage{float}

\begin{document}
\title{Ensemble Modeling for Multimodal Visual Action Recognition}
%
%
\author{Jyoti Kini \and
Sarah Fleischer \and
Ishan Dave \and
Mubarak Shah}
%
%
\institute{Center for Research in Computer Vision, University of Central Florida, USA 
\email{\{jyoti.kini, sa420832, ishandave\}@ucf.edu, shah@crcv.ucf.edu}}
\maketitle              
\begin{abstract}
In this work, we propose an ensemble modeling approach for multimodal action recognition. We independently train individual modality models using a variant of focal loss tailored to handle the long-tailed distribution of the MECCANO \cite{ragusa2021meccano} dataset. Based on the underlying principle of focal loss, which captures the relationship between tail (scarce) classes and their prediction difficulties, we propose an exponentially decaying variant of focal loss for our current task. It initially emphasizes learning from the hard misclassified examples and gradually adapts to the entire range of examples in the dataset. This annealing process encourages the model to strike a balance between focusing on the sparse set of hard samples, while still leveraging the information provided by the easier ones. Additionally, we opt for the {\it{late fusion}} strategy to combine the resultant probability distributions from RGB and Depth modalities for final action prediction. Experimental evaluations on the MECCANO dataset demonstrate the effectiveness of our approach.
\end{abstract}

\section{Introduction}
Amidst the surge of data in recent times, multimodal learning has emerged as a transformative approach, leveraging heterogeneous cues from multiple sensors to enhance the learning process. Both early and late multimodal fusion mechanisms have demonstrated the ability to effectively harness complementary information from diverse sources. However, real-world multimodal data associated with action occurrences suffer from an inherent skewness, giving rise to the long-tailed action recognition scenario. In such cases, some action classes are prevalent and well-represented in the training data, while others are scarce, leading to significant data imbalance. The inherent complexity of multimodal data, combined with such data imbalance, presents a formidable challenge for learning approaches.

In order to fuse information from different data streams, researchers in the vision community have proposed a variety of approaches, spanning from consolidating feature representations at an early stage (early fusion) \cite{nishida2016multimodal,landi2019perceive} to aggregating prediction scores at the final stage (late fusion) \cite{ding2015robust,koo2018cnn}. Furthermore, to combat the long-tailed visual recognition issue, use of data augmentation \cite{wang2021rsg,verma2019manifold,zhang2017mixup,xiang2021increasing}, re-sampling \cite{more2016survey,kim2020adjusting,buda2018systematic}, cost-sensitive loss \cite{lin2017focal,huang2016learning,cui2019class,wang2021contrastive}, and transfer learning \cite{samuel2021generalized,liu2019large,zhou2020bbn,zhong2019unequal,li2021self,jamal2020rethinking} strategies is highly recommended. Methods that rely on data augmentation techniques like M2M \cite{kim2020m2m}, ImbalanceCycleGAN\cite{sahoo2020mitigating}, and MetaS-Aug\cite{li2021metasaug} attempt to augment the minority classes with diverse samples. Other data augmentation-based works \cite{liu2021semi,yang2020rethinking,wei2021crest} generate pseudo labels to reduce scarcity in tail classes. However, these methods are often limited by their ability to generate realistic and diverse minority class samples. Some of the re-sampling approaches \cite{peng2019trainable,ren2020balanced,kang2019decoupling,buda2018systematic} focus on assigning larger sampling probabilities to the tail classes. Although beneficial, re-sampling models are at risk of overfitting to the tail classes.

In this paper, we introduce an ensemble training strategy that leverages multimodal RGB and Depth signals for visual action recognition, using the late fusion mechanism. We, also, allow the model to focus on {\emph hard-to-classify examples} using an exponentially decaying variant of the focal loss objective function. This function not only reduces the KL divergence between the predicted distribution and the ground-truth distribution but also simultaneously increases the entropy of the predicted distribution, thereby preventing model overconfidence towards majority classes. 

\begin{figure}[t]
  \centering
  \includegraphics[width=\textwidth]{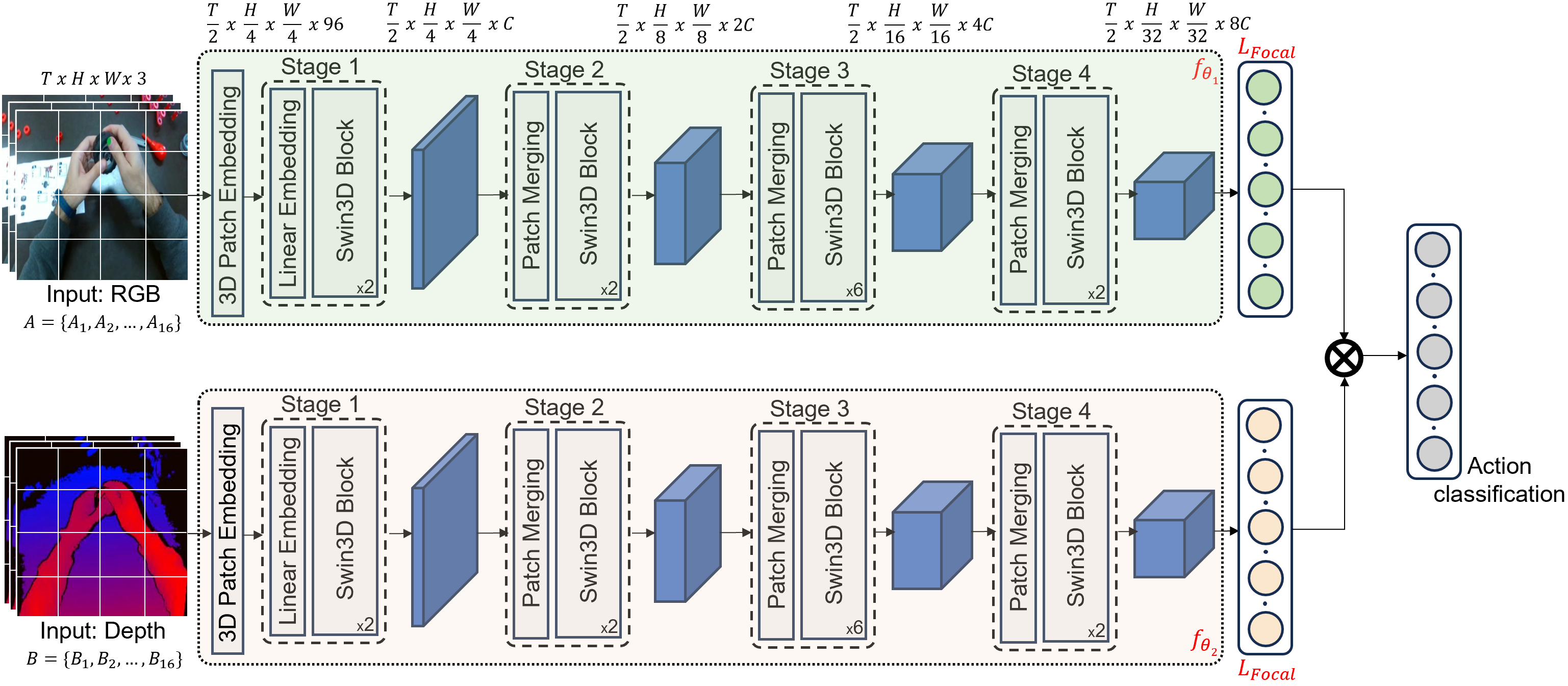}
  \caption{\textbf{Architecture}: The RGB frames $\{A_{i}$, $A_{i-1}$,..,$A_{T}\}$ and Depth frames $\{B_{i}$, $B_{i-1}$,..,$B_{T}\}$ are passed through two independently trained Swin3D-B \cite{liu2022video} encoders ${f}_{\theta_1}$ and ${f}_{\theta_2}$ respectively to generate feature tokens. The resultant class probabilities, obtained from each pathway, are averaged to subsequently yield action classes. Exponentially decaying focal loss $L_{Focal}$ is leveraged to deal with the long-tailed distribution exhibited by the data.} 
  \label{fig:Architecture}
\end{figure} 

\section{Approach}
\subsection{Cross-Modal Fusion}
\label{Cross-Modal Fusion}
Figure \ref{fig:Architecture} provides comprehensive details of our proposed approach.
Given a set of spatiotemporally aligned RBG and Depth sequences that extend between $[t_s, t_e ]$, where $t_s$ and $t_e$ are the start and the end duration of the sequence, our goal is to predict the action class $\mathcal{O} = \{o_{1}$, $o_{2}$,.., $o_{K}\}$ associated with the sequence. In order to achieve this, we adopt an ensemble architecture comprising two dedicated Video Swin
Transformer \cite{liu2022video} backbones to process the RGB clip $\mathcal{A} = \{A_{i}$, $A_{i-1}$,..,$A_{T}\}$ and Depth clip $\mathcal{B} = \{B_{i}$, $B_{i-1}$,..,$B_{T}\}$ independently. Here, $i$ corresponds to a random index spanning between $t_s$ and $t_e$. The input video for each modality defined by size $T\times H\times W\times 3$ results in token embeddings of dimension $\frac{T}{2}\times H_d\times W_d\times C$. We pass this representation retrieved from {\it{stage-4}} of the base feature network to our newly added fully connected layer and fine-tune the overall network. The final prediction is derived by averaging the two probability distributions obtained as output from the RGB and Depth pathways. 

\subsection{Exponentially Decaying Focal Loss}
\label{Losses}
Focal loss \cite{lin2017focal} is a variant of cross-entropy loss with a modulating factor that down-weighs the impact of easy examples and focuses on the hard ones. It, therefore, tends to prevent bias towards data-rich classes and improves the performance on scarce categories. \\
Multi-classification cross-entropy (CE) loss is given by:
\begin{align}
    L_{CE} = - \sum_{j=1}^{K}{y_j \log(p_j)} 
\end{align}
where, say we have $K$ action classes, and $y_j$ and $p_j$ correspond to the ground-truth label and predicted probability respectively for the $j^{th}$ class. \\
On the other hand, the key objective of focal loss \cite{lin2017focal} is defined as:
\begin{align}
    L_{Focal} = - \sum_{j=1}^{K} {(1-p_j)^{\gamma} \log p_j}
\end{align}
In our work, we use focal loss $L_{Focal}$  and exponentially decay $\gamma$ from $2$ to $0.1$. When $\gamma$=0, the objective function is equivalent to cross-entropy loss. Our proposed annealing process for $\gamma$ allows for the model to focus on the sparse set of hard examples in the early stage of training, and gradually shift its focus towards easy examples. This configuration is essential to ensure that the model learns meaningful representations and generalized decision boundaries.

\section{Experimental Setup}
\subsection{Data-preprocessing}
For our experiments, we resize the frames to a width of $256$, without disturbing the aspect ratio of the original image, followed by a random crop of $224 \times 224$. In addition, we use 16 consecutive frames to generate a single clip for the forward pass. In the case of shorter sequences, we pad the sequence with the last frame.

\subsection{Training}
We use the Swin3D-B \cite{liu2022video} backbone, which is pre-trained on the Something-Something v2 \cite{mahdisoltani2018effectiveness} dataset. We adopt focal loss \cite{lin2017focal} with exponentially decaying $\gamma$ for training the classification model. For optimization, AdamW optimizer with a learning rate of $3 \times 10^{-4}$ and a weight decay of 0.05 has been employed. Our model converges in about 20 epochs on the MECCANO dataset. We report the {\it{Top-1}} and {\it{Top-5}} classification accuracy as our evaluation metrics. Additionally, to demonstrate the effectiveness of employing the focal loss for this task, we present the average class {\it{Precision}}, {\it{Recall}} and {\it{F1-score}}.

\setlength{\tabcolsep}{4pt}
\begin{table*}[!th]
\begin{center}
\renewcommand{\arraystretch}{1}
\begin{tabularx}{10.5cm}{l*{13}{c}{c}}
\toprule
{Modality} & Loss & \multicolumn{2}{c}{Accuracy} & \multicolumn{2}{c}{AVG Class} & AVG {\it F1-score}  \\
\cmidrule(lr){3-4}                  
\cmidrule(lr){5-6}
& & {\it Top-1} & {\it Top-5} & {\it Precision} & {\it Recall} & \\
\midrule
RGB & {\it{CE}} & 48.35 & 80.91 & 45.52 & 48.35 & 46.22 \\
Depth & {\it{CE}} & 43.32 & 75.38 & 41.79 & 43.32 & 41.88 \\
RGB+Depth & {\it{CE}} & 50.94 & 81.79 & 47.28 & 50.94 & 48.08 \\
\midrule
RGB & Focal & 50.80 & 82.36 & 47.17 & 50.80 & 47.95 \\
Depth & Focal & 45.52 & 78.07 & 43.74 & 45.52 & 43.41 \\
RGB+Depth & Focal & 52.82 & 83.85 & 49.97 & 52.82 & 49.41 \\
\midrule
RGB$^*$ & Focal & 53.03 & 85.37 & 50.46 & 53.03 & 50.39 \\
Depth$^*$ & Focal & 48.39 & 80.55 & 46.43 & 48.39 & 46.35 \\
RGB+Depth$^*$ & Focal & {\bf 55.37} & 85.58 & 52.41 & 55.37 & 52.28 \\
\bottomrule
\end{tabularx}
\end{center}
\caption{Results demonstrating the effectiveness of our ensemble modeling approach for the action recognition task on the MECCANO test dataset. {\it{CE}} implies Cross-Entropy loss. {\bf *} refers to model trained using both train$+$validation set.}
\label{table:Results}
\end{table*}
 
\section{Discussion}
\begin{figure}[H]
\begin{center}
   \includegraphics[width=0.5\linewidth]{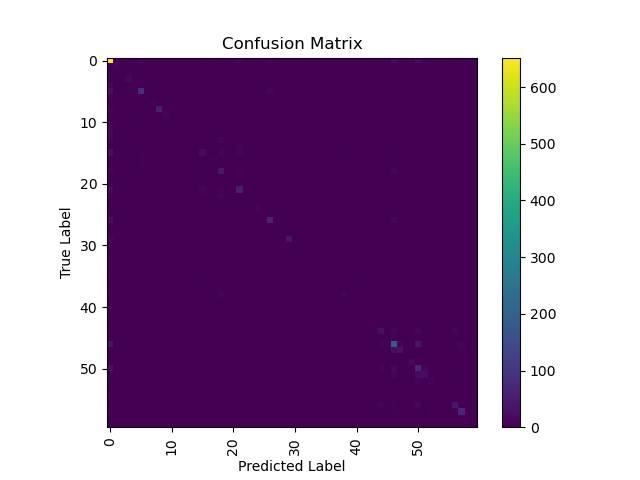}
\end{center}
   \caption{The resultant confusion matrix obtained from the MECCANO test set highlights our model's proficiency in handling the long-tailed distribution.}
\label{fig:Confusion}
\end{figure}

Table \ref{table:Results} presents our results on the MECCANO test set. Applying cross-entropy loss to fine-tune our model, pre-trained on Something-Something v2, gives us an initial baseline accuracy of 50.94\% on our multimodal setup. Introducing focal loss with exponential decay in $\gamma$ boosts the overall accuracy by $\approx$ 2\%. Figure \ref{fig:Confusion} demonstrates the effectiveness of our approach in dealing with the long-tailed distribution of the MECCANO dataset. Furthermore, combining the train and validation data gives the best {\it{Top-1}} accuracy of 55.37\%. 

\newpage

\bibliographystyle{splncs04}
\bibliography{egbib}
\end{document}